\documentclass[letterpaper,twocolumn,10pt]{article}
\usepackage{usenix2019_v3}
\usepackage{tikz}
\usepackage{amsmath}
\usepackage{mathptmx}
\usepackage{indentfirst}
\usepackage{balance}
\usepackage{filecontents}
\begin{document}

\date{}
\title{\Large \bf A Novel Approach to Guard from Adversarial Attacks using Stable Diffusion}

\author{
{\rm Uma Maheswara Rao Meleti}\\
Clemson University
\and
{\rm Trinath Sai Subhash Reddy Pittala}\\
Clemson University
\and
{\rm Geethakrishna Puligundla}\\
Clemson University
}

\maketitle
\begin{abstract}
Recent developments in adversarial machine learning have highlighted the importance of building robust AI systems to protect against increasingly sophisticated attacks. While frameworks like AI Guardian are designed to defend against these threats, they often rely on assumptions that can limit their effectiveness. For example, they may assume attacks only come from one direction or include adversarial images in their training data. Our proposal suggests a different approach to the AI Guardian framework. Instead of including adversarial examples in the training process, we propose training the AI system without them. This aims to create a system that is inherently resilient to a wider range of attacks. Our method focuses on a dynamic defense strategy using stable diffusion that learns continuously and models threats comprehensively. We believe this approach can lead to a more generalized and robust defense against adversarial attacks.

In this paper, we outline our proposed approach, including the theoretical basis, experimental design, and expected impact on improving AI security against adversarial threats.
\end{abstract}

\section{Introduction}

Artificial Intelligence (AI) systems are increasingly integral to various applications, from autonomous vehicles to cybersecurity. However, the rise of adversarial attacks—manipulations intended to deceive AI models—poses significant challenges to the reliability and safety of these systems. The AI Guardian framework represents a significant stride in adversarial defense, employing innovative strategies to detect and mitigate such attacks. It operates primarily assuming that attacks can be anticipated in a specific direction, utilizing known vulnerabilities as backdoors for protective measures.

While the AI Guardian model has made significant contributions, it has limitations that could affect its effectiveness in real-world situations. One such limitation is its reliance on including adversarial examples in training datasets. These examples may not accurately reflect the evolving tactics of adversaries. Additionally, the framework assumes that attacks follow a predictable, unidirectional pattern, which is not true with modern cyber threats' complex and multi-directional nature.

This paper explores these limitations while proposing an alternative approach that seeks to enhance the robustness and adaptability of the AI Guardian. By adopting a holistic view of adversarial contexts and reframing the defensive mechanisms accordingly, our approach intends to establish a more dynamic and resilient framework using stable diffusion. Through a comprehensive analysis of existing methodologies and the integration of advanced, adaptive defensive strategies, we aim to significantly reduce vulnerabilities in AI systems and improve their ability to respond to unpredictable adversarial challenges.

\section{Related Work to AI Guardian}
\subsection{Trapdoor Defense Mechanism} 

The Trapdoor defense mechanism is an advanced approach that employs backdoors to counter adversarial attacks. It inserts multiple class-agnostic backdoors into the model, each targeting a specific label, acting as bait to trap adversarial examples. By comparing inputs to embedded triggers, Trapdoor can detect adversarial examples. However, embedding too many backdoors may impact its effectiveness, as it was designed to protect up to 100 labels, leaving others vulnerable. Additionally, Trapdoor has shown susceptibility to adaptive attacks, like Feature-indistinguishable attacks. In contrast, AI-Guardian only requires one backdoor with a single trigger, making it scalable and efficient in safeguarding multi-label models.

\subsection{Online Defense Mechanisms Against Adversarial Attacks}

Online defense mechanisms are deployed during the model prediction phase to counteract adversarial attacks in real-time. These solutions are crucial for maintaining model integrity and performance during inference. Key strategies include:
\begin{itemize}
    \item \textbf{Input Reconstruction}: Aims to restore inputs to their original, unaltered state.
    \item \textbf{Real-Time Anomaly Detection}: Identifies and mitigates suspicious input patterns.
    \item \textbf{Adaptive Model Retraining}: Adjusts the model dynamically based on detected adversarial patterns.
\end{itemize}
Each strategy focuses on minimizing the impact of attacks, ensuring robust and reliable model predictions under adversarial conditions.

\subsubsection{JPEG Compression and Total Variance Minimization}

In adversarial defense, preprocessing techniques play a crucial role in preparing input images before they are fed into the model. Methods like JPEG Compression and Total Variance Minimization are particularly important as they focus on reducing the impact of perturbations introduced by attackers. These techniques help in making the input images more robust against adversarial attacks, thereby enhancing the overall security of the model:

\begin{itemize}
    \item \textbf{JPEG Compression}: This technique leverages the JPEG image compression algorithm, which inherently reduces high-frequency content in the image data. By doing so, it effectively diminishes subtle adversarial perturbations, which are often high-frequency elements that are not readily perceptible to the human eye.
    
    \item \textbf{Total Variance Minimization}: This method focuses on reducing the total variance within the image. It smooths the image by minimizing the sum of the squared differences between adjacent pixel values, thereby blurring out and reducing the impact of minor, noise-like perturbations that could mislead the model.
\end{itemize}

Both techniques are designed to simplify the image data, thus reducing the likelihood that minor, calculated modifications by an attacker will remain effective after preprocessing. This ensures that a cleaner, more robust input is fed into the model, enhancing overall security against adversarial attack.

\subsubsection{DISCO and Denoiser}

Within the framework of adversarial defense, two prominent techniques are DISCO and various denoising methods. Each serves to refine input data before it reaches the model, aiming to mitigate the effects of adversarial manipulations:

\begin{itemize}
    \item \textbf{DISCO}: This technique applies localized manifold projections to filter out perturbations in the data effectively. By projecting input data onto the expected data manifold, DISCO can significantly reduce deviations introduced by adversarial attacks, ensuring that the inputs adhere more closely to the distribution of clean data.
    
    \item \textbf{Denoiser}: Denoising methods encompass a range of techniques designed to cleanse input images from noise-like perturbations. These can include statistical methods, neural networks, or signal processing techniques that target and eliminate the adversarial noise, thereby purifying the input and bolstering the model’s resilience to attacks.
\end{itemize}

Both DISCO and denoising techniques are crucial for enhancing the security of machine learning models by ensuring that the processed inputs are devoid of malicious modifications that could otherwise compromise model accuracy.

\subsubsection{Latent Intrinsic Dimensionality and Argos}

The methodologies of Latent Intrinsic Dimensionality (LID) and Argos are instrumental in identifying and countering adversarial attacks on machine learning models, each focusing on distinct aspects of input analysis:

\begin{itemize}
    \item \textbf{Latent Intrinsic Dimensionality (LID)}: LID measures the internal characteristics of data representations within a model, exploiting differences that typically manifest between normal and adversarial inputs. By evaluating how these characteristics diverge from expected norms, LID can help in detecting inputs that have been subtly manipulated to deceive the model.

    \item \textbf{Argos}: This approach leverages regeneration mechanisms that reconstruct input images based on their predicted labels. Argos evaluates the fidelity of these regenerated images against the original inputs, effectively detecting adversarial examples by spotting significant deviations. This technique helps to ensure that only inputs which maintain consistency with their expected characteristics under the model's logic are processed.
\end{itemize}

Together, LID and Argos provide a comprehensive shield against adversarial attacks, enhancing the robustness of AI systems by improving their ability to discern and reject manipulated inputs.

\subsubsection{Morphence}

Morphence introduces a novel approach to enhancing the security of machine learning systems against adversarial attacks by leveraging a pool of models. This method involves:

\begin{itemize}
    \item \textbf{Model Pooling}: Morphence creates a diverse pool of models, each trained with slightly different data sets or hyperparameters. This diversity ensures that the internal workings and decision boundaries of each model vary.
    
    \item \textbf{Random Model Selection}: When processing queries, Morphence randomly selects a model from the pool to handle each request. This randomness introduces unpredictability in model predictions, making it significantly more difficult for attackers to tailor perturbations that consistently fool the system across all models.
\end{itemize}

By integrating variability and unpredictability into the model prediction phase, Morphence effectively counters the precise manipulations characteristic of adversarial attacks, thus safeguarding the integrity of AI applications.

\subsection{Adversarial Training Approaches} 

Adversarial training is a common offline defense mechanism in adversarial machine learning, where the training dataset is augmented with adversarial examples. These examples, crafted by adding subtle perturbations to clean input data, aim to trick the model into making incorrect predictions. By training with clean and adversarial examples and their accurate labels, the model improves its ability to generalize and resist adversarial attacks. Although adversarial training enhances model resilience, it cannot wholly eliminate vulnerabilities to such attacks and only reduces the success rate to a certain degree. Conversely, AI-Guardian handles adversarially altered inputs more effectively and offers a more cost-efficient solution.\\

In the realm of AI security, significant work has been
done to address the susceptibility of neural networks to adversarial attacks. Key approaches include:

\begin{itemize}
    \item \textbf{JPEG Compression and Total Variance Minimization:} These methods compress the input images before feeding them into the model to eliminate the perturbations added by the attackers.
    
    \item \textbf{DISCO and Denoisers:} These techniques adopt localized manifold projections to remove the perturbations of adversarial attacks. A Denoiser can also be used to remove the perturbations on input images to defeat adversarial attacks.
    
    Giving wrong prediction results deliberately or refusing to provide prediction results directly features squeezing smooths input images fed into the model and tries to detect adversarial examples by computing the distance between the prediction vectors of the original inputs and the squeezed inputs.
    
    \item \textbf{Feature Squeezing:} It simplifies the inputs to reduce the effect of minor perturbations.
    
    \item \textbf{Argos} uses a set of regeneration mechanisms to reproduce the input image according to the label predicted by the model and detects the adversarial example by checking whether the reproduced image deviates significantly from the original.
    
    \item \textbf{Morphence} creates a pool of models and uses different models to process different queries to introduce randomness in prediction and defeat adversarial attacks.
    
    \item \textbf{Trapdoor} embeds multiple backdoors into the to-be-protected model. Each backdoor is class-agnostic, targeting a specific label.
\end{itemize}
Each of these methods has shown varying degrees of success in mitigating the impact of adversarial attacks, with a focus on balancing the robustness of the network against the fidelity of its predictions. The ongoing research in this field reflects the complexity and evolving nature of AI security challenges.

\section{Method}

The crux of the paper is Batch learning of training with triggers and noise applied to initial points.
The following algorithm is for training:

    \begin{figure}[htbp]
      \centering
      \includegraphics[width=0.5\textwidth]{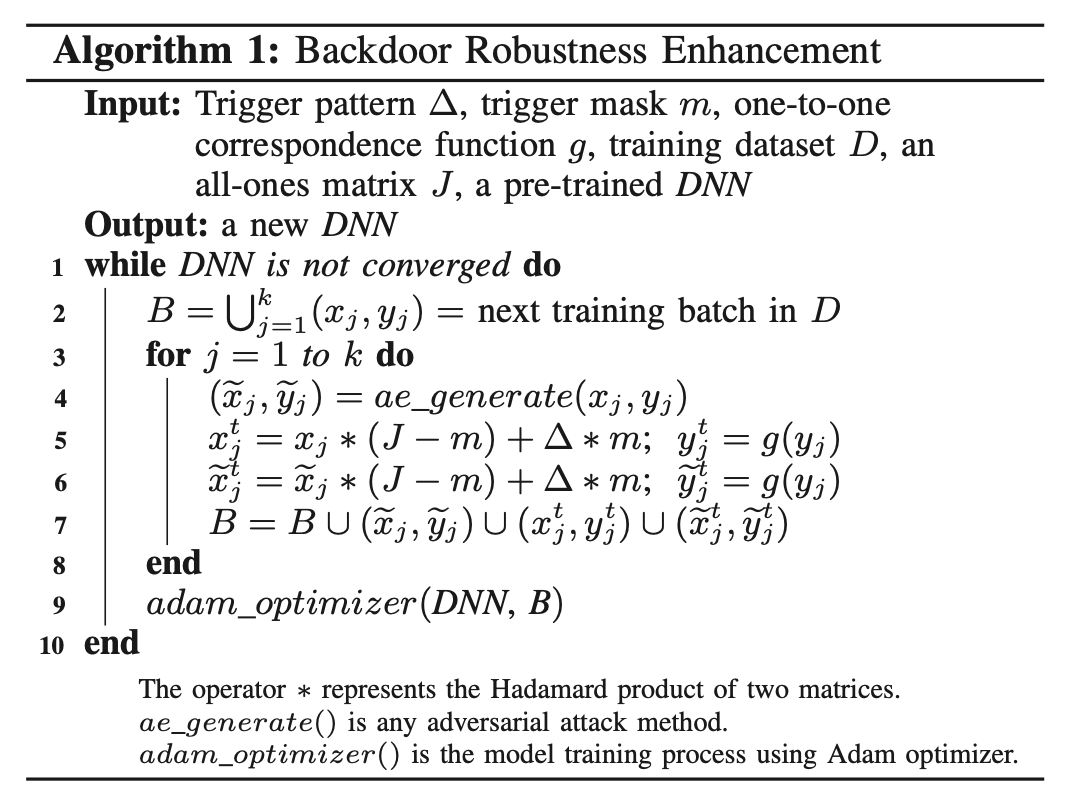}
      \caption{Backdoor Robustness Enhancement}
      \label{fig:my_label}
    \end{figure}

    \begin{figure}[htbp]
      \centering
      \includegraphics[width=0.5\textwidth]{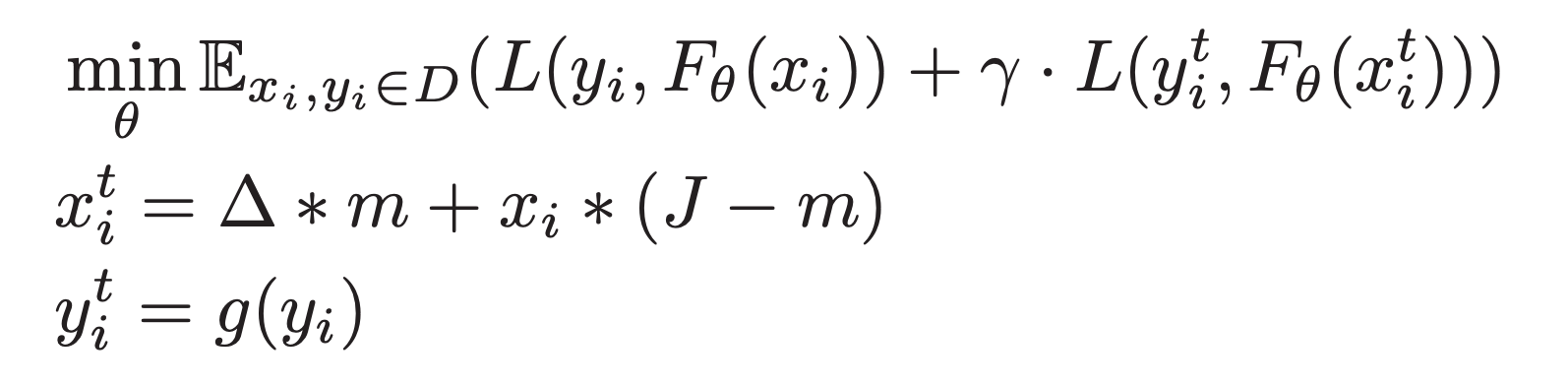}
      \caption{It is the loss function for the Backdoor Algorithm. The hyperparameter is set to change dynamically after a few batches depending on trigger data and original data resistance to noise.}
      \label{fig:my_label}
    \end{figure}

\section{Evaluation and Reproduced Results}

In our investigation, we carefully examined the methodologies outlined in the original research paper and attempted to replicate the findings. Our experimental reproduction involved applying the described method to the MNIST dataset under both normal conditions and in the presence of a specific adversarial threat known as the Carlini Wagner attack. The results were markedly different:\\ 
\textbf{AI Guardian} \cite{aiGuardian} demonstrated a high success rate of 0.94 in a benign environment but dropped drastically to 0.01 when subjected to an adversarial attack. These results suggest that while the model performs well under normal conditions, its ability to defend against advanced attacks requires substantial improvement.

    \begin{figure}[htbp]
      \centering
      \includegraphics[width=0.45\textwidth]{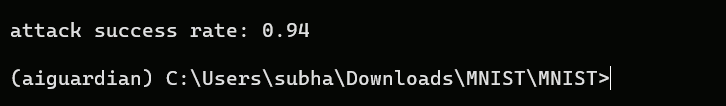}
      \caption{Accuracy without Defense}
      \label{fig:my_label}
    \end{figure}

    \begin{figure}[htbp]
      \centering
      \includegraphics[width=0.45\textwidth]{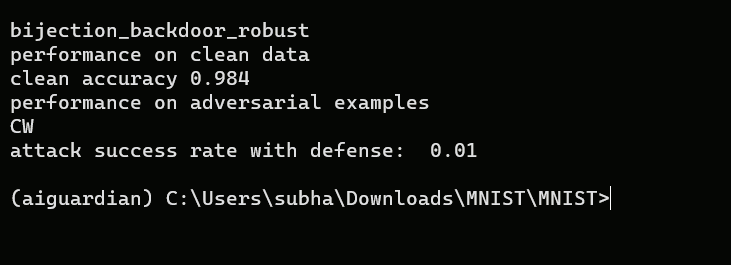}
      \caption{Accuracy with Defense}
      \label{fig:my_label}
    \end{figure}

We tested the other 3 datasets (GTSRB, VGG-Face, Youtube-Face) on a Palmetto Node. The MINST test was done on the local machine \& results are shown above. For all the datasets, we observed the same results mentioned in the original paper.

\section{Limitations of AI-Guardian}

AI-Guardian is predicated on the belief that attacks will be unidirectional. This assumption is problematic because attacks can manifest in diverse and unexpected ways. Relying exclusively on this assumption can expose the model to various vulnerabilities that cannot be mitigated by a single "back door" defense mechanism.

The AI-Guardian framework integrates adversarial and triggered images into training datasets to enhance models' resilience against attacks. 
When both adversarial and triggered images are included, the process essentially reduces to dictionary mapping. This becomes a dictionary mapping for attack recognition and mitigation, which may lead to unreliable experimental results and limit the method's effectiveness against sophisticated attacks.

It also assumes that adversarial tactics are static, which is unrealistic; attackers constantly evolve and adapt their strategies. Therefore, training with a fixed set of adversarial images may not offer a robust defense against newly developed attack vectors.

\section{Proposed Approach}

We propose a new approach using generative AI to defend against adversarial attacks. Our approach essentially uses stable diffusion to refine adversarial attacks, which we think is more futuristic and can defend both white box and black box attacks. Stable diffusion predicts noise from the input image and denoises it back to the input image, which makes it difficult for attackers to craft attacks to manipulate noise for a series of time steps.


\subsection{Stable Diffusion}

Stable Diffusion's \cite{stableDiffusion} unconditional image-to-image translation capabilities leverage its diffusion-based architecture to transform input images in powerful ways. The process begins by encoding the input image into a compact latent representation, then iteratively denoised through a diffusion process to generate a new latent space. This transformed latent is finally decoded back into an output image that shares similarities with the original but exhibits novel variations and alterations guided by the model's learned representations. This versatile image-to-image translation allows Stable Diffusion to be used for tasks like image manipulation, artistic expression, and visual prototyping, where users can explore a wide range of transformations starting from an existing image.

    \begin{figure}[htbp]
      \centering
      \includegraphics[width=0.45\textwidth]{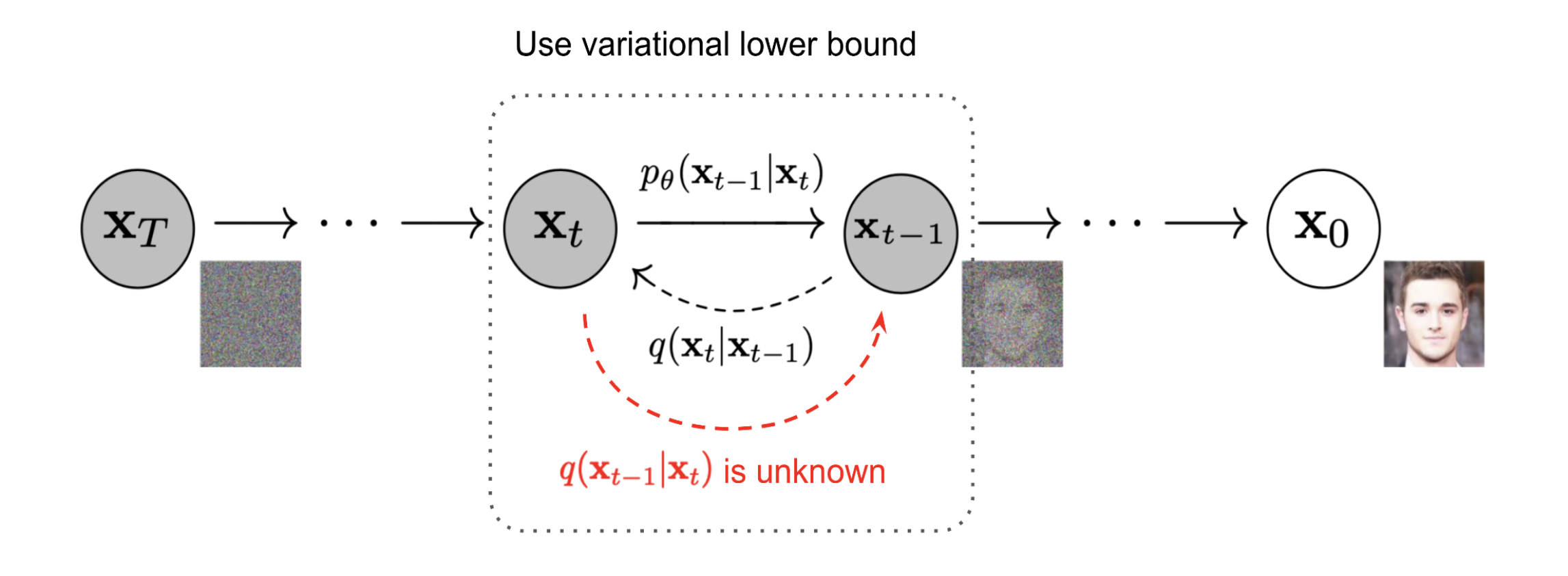}
      \caption{Stable Diffusion}
      \label{fig:sd}
    \end{figure}

    \begin{figure*}[htbp]
      \centering
      \includegraphics[width=1\textwidth]{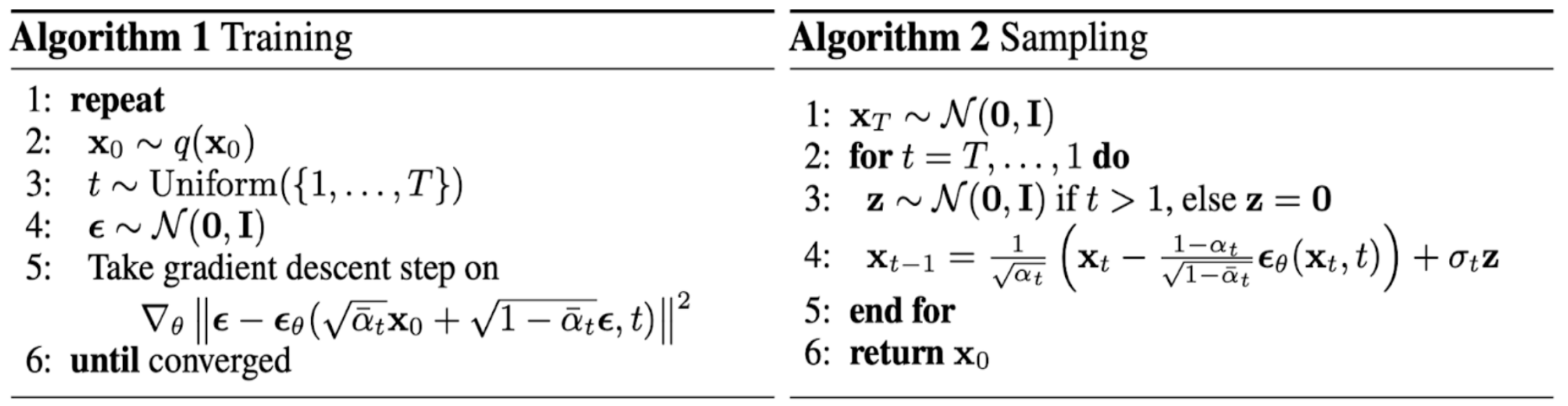}
      \caption{Stable Diffusion}
      \label{fig:sda}
    \end{figure*}

\subsection{Training}

Stable Diffusion is a large text to image diffusion model trained on billions of images. Image diffusion model learn to denoise images to generate output images. Stable Diffusion uses latent images encoded from training data as input. Further, given an image $zo$, the diffusion algorithm progressively add noise to the image and produces a noisy image $zt$, with $t$ being how many times noise is added. When $t$ is large enough, the image approximates pure noise. Given a set of inputs such as time step t, text prompt, image diffusion algorithms learn a network to predict the noise added to the noisy image $zt$.

\subsection{Sampling}

Image generation with Stable Diffusion involves a denoising process that converts random noise into a representative image sample without any additional guidance or conditioning. The process involves generating a random noise image as the starting point for the denoising process. Iteratively applying the denoising model, which predicts the noise residual at each step, to gradually remove noise and refine the image. The number of denoising steps controls the quality of the final image - more steps typically produce higher quality but take longer.
The denoised image is the final output, which resembles a random sample from the training data distribution that the model was trained on.

\subsection{Algorithm: Stable Diffusion Training Loop}

\begin{verbatim}
Input: Training dataset D, adversarial 
perturbation function perturb(), 
number of epochs E
Output: Trained Model M

1: Initialize Model M
2: for each epoch in E do
3:     for each batch B in D do
4:         Generate adversarial examples A using 
           perturb(B)
5:         Combine B and A to form new training
           batch N
6:         Train M on N using standard loss function
7:     end for
8: end for
9: return M
\end{verbatim}

\subsection{Algorithm: Stable Diffusion Sampling Loop}
\begin{verbatim}
Input: Initial input seed S, model M, number of 
iterations I
Output: Generated output O

1: Set initial input X = S
2: for i from 1 to I do
3:     Apply model M to X to generate new input X'
4:     Update X to X'
5: end for
6: Set O to final state of X
7: return O
\end{verbatim}

\subsection{White Box Attack}
\begin{figure}[htbp]
  \centering
  \includegraphics[width=0.45\textwidth]{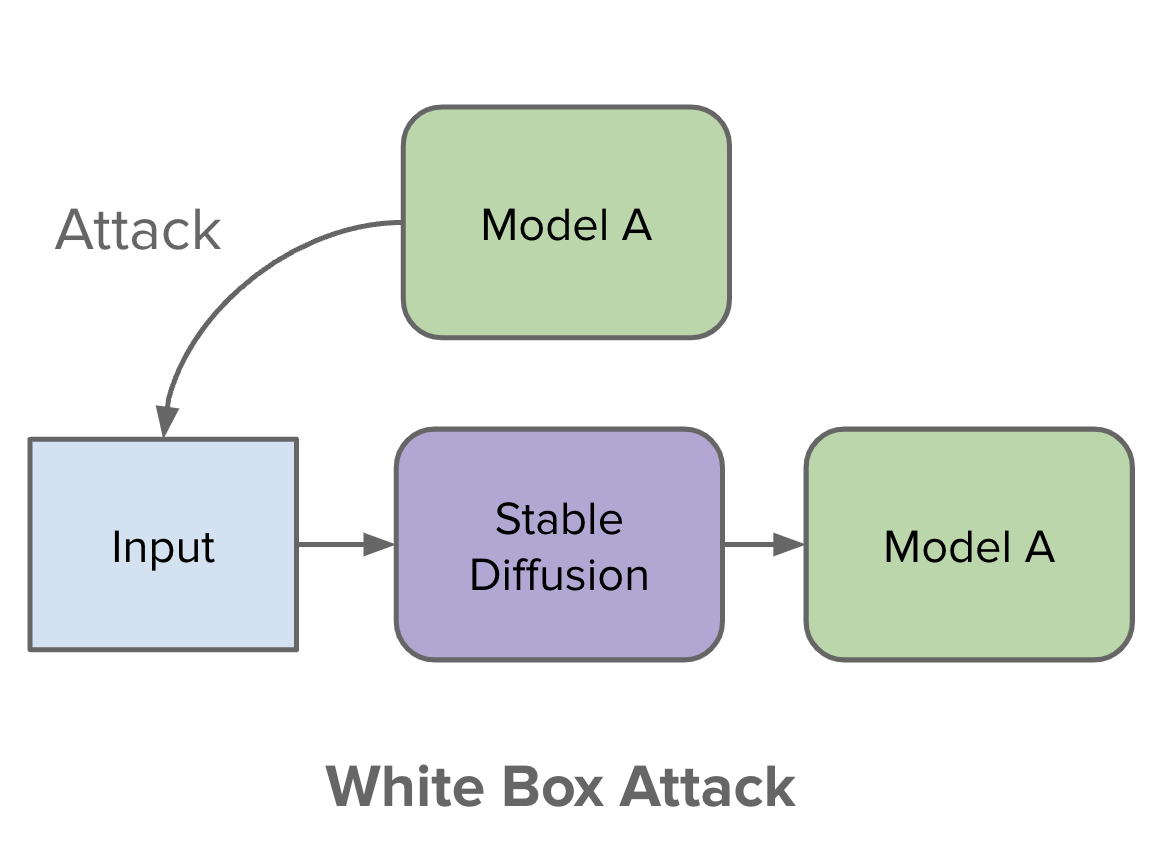}
  \caption{Illustration of a White Box Attack on Stable Diffusion Models}
  \label{fig:wba}
\end{figure}

A white box attack is an attack on a deep learning model, where the same model gradients are used to attack the image to perturb the input image so that it can be classified into a different class in the context of image classification. We simulated the white box scenario by training classifier A and using the same classifier to attack the images. We saw both the performances of the attacked images and refined attacked images using stable diffusion on classifier A, which was the same classifier used for attacking. It is illustrated in \ref{fig:wba}

\subsection{Black Box Attack}
\begin{figure}[htbp]
  \centering
  \includegraphics[width=0.45\textwidth]{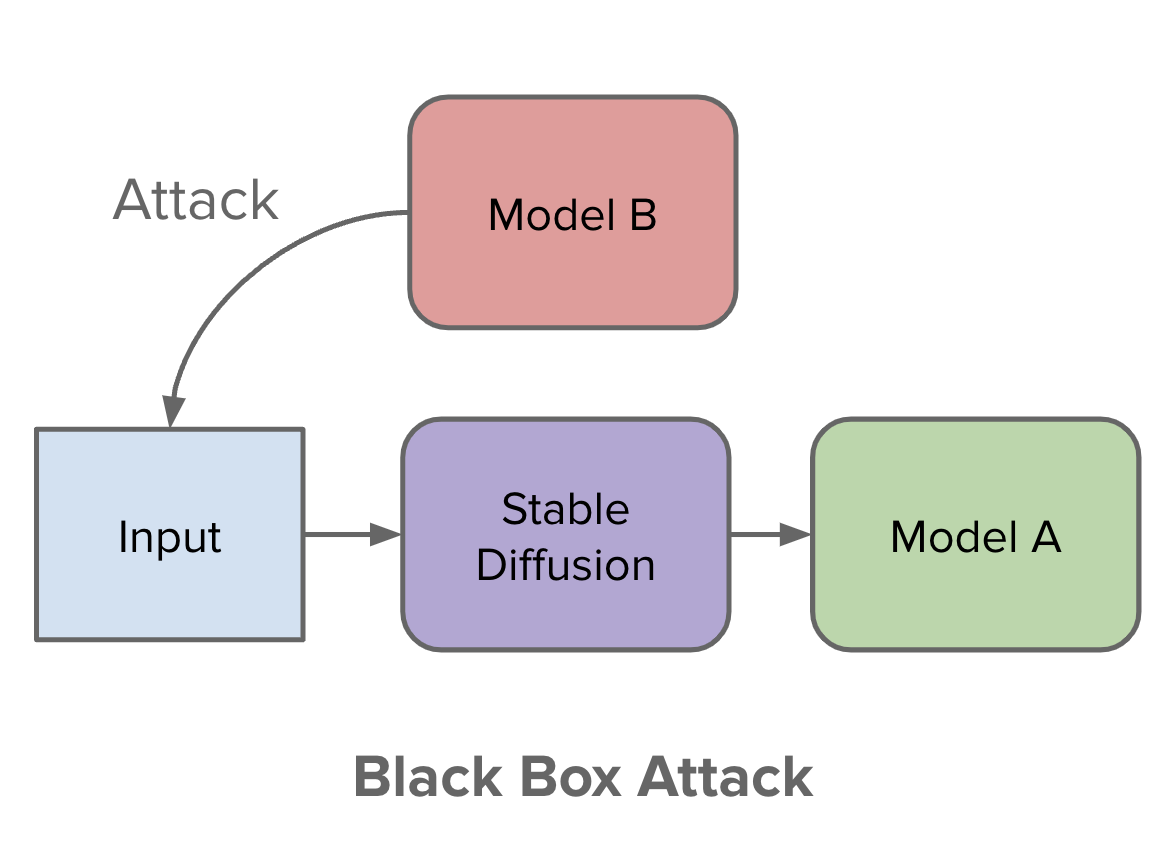}
  \caption{Illustration of a White Box Attack on Stable Diffusion Models}
  \label{fig:bba}
\end{figure}

In the black box attack, we simulate a different scenario where we assume the attacker uses an unknown model to attack the classifier. We trained classifier B, which will be used to attack images fed to previously trained classifier A. This represents unknown attacks in various directions that are more aligned with real-time attacks. We inferred both attacked images and refined images with stable diffusion and saw an increase in performance in the latter case. It is illustrated in \ref{fig:bba}

\subsection{Attack Methods}
\subsubsection{Projected Gradient Descent (PGD)}

Projected Gradient Descent (PGD) is an iterative attack method used in adversarial machine learning to generate adversarial examples. Recognized for its effectiveness and flexibility, PGD is often termed as a "universal" adversarial attack. The method involves taking multiple small gradient steps and then projecting these steps onto an epsilon-ball around the original input. This ensures the perturbations remain within the limits of human imperceptibility.

\textbf{Algorithm Overview:}
\begin{itemize}
    \item \textbf{Initialization:} Start with an input that is already slightly perturbed.
    \item \textbf{Iterative Update:} For each iteration, adjust the input by moving in the direction that maximizes the loss of the model, constrained by the perturbation's magnitude.
    \item \textbf{Projection:} After each gradient step, project back the perturbed example onto the epsilon-ball around the original input to ensure that the perturbation does not exceed the defined threshold.
\end{itemize}

\textbf{Algorithm:}
\begin{verbatim}
Input: Original input X, target label Y, model M, 
loss function L, step size alpha, perturbation 
limit epsilon, number of iterations N
Output: Adversarial example X_adv

1: Initialize X_adv = X + RandomNoise() 
   within epsilon limit
2: for i from 1 to N do
3:     Compute gradient: G = Gradient
       of L with respect to X_adv for Y
4:     Update adversarial example:
        X_adv = X_adv + alpha * Sign(G)
5:     Project X_adv back into the 
       epsilon-ball around X:
        X_adv = X
            + Clip(X_adv - X, -epsilon, epsilon)
6: end for
7: return X_adv
\end{verbatim}

\textbf{Significance:}
PGD is widely used both as an attack to test the robustness of neural networks and as a technique for adversarial training, which enhances model robustness by incorporating adversarial examples into the training process. The iterative nature of PGD allows it to find more effective adversarial examples compared to simpler, one-step methods.

\subsubsection{Fast Gradient Sign Method (FGSM)}

The Fast Gradient Sign Method (FGSM) is a simple yet powerful technique to generate adversarial examples, which are inputs crafted to cause a machine learning model to make a mistake. Developed by Ian Goodfellow and colleagues, FGSM works by using the gradients of the neural network to create perturbations that maximize the loss of the model. These perturbations are added to the original input to mislead the model.

\textbf{Algorithm Overview:}
\begin{itemize}
    \item \textbf{Compute Gradients:} Calculate the gradients of the loss function with respect to the input data.
    \item \textbf{Create Perturbation:} Apply the sign of these gradients to the inputs, which helps in determining the direction to alter the input to maximize the loss.
    \item \textbf{Apply Perturbation:} Adjust the original input by a small factor epsilon in the direction of the gradient sign. This operation ensures the changes are minimal but sufficient to deceive the model.
\end{itemize}

\textbf{Algorithm:}
\begin{verbatim}
Input: Original input X, target label Y, model M,
loss function L, perturbation magnitude epsilon
Output: Adversarial example X_adv

1: Compute the gradient of the loss L w.r.t X:
   G = Gradient of L(X, Y) with respect to X
2: Create perturbation:
   Perturbation = epsilon * Sign(G)
3: Generate adversarial example:
   X_adv = X + Perturbation
4: return X_adv
\end{verbatim}

\textbf{Significance:}
FGSM is favored for its computational efficiency, making it particularly useful for benchmarking the robustness of models in a fast and straightforward manner. However, due to its simplicity, FGSM often finds suboptimal adversarial examples compared to more sophisticated methods like PGD. Nevertheless, it remains a popular choice for initial adversarial robustness checks and introductory studies into adversarial machine learning.

\section{Results}

\subsection{White Box Attack - PGD}
This section presents visual results from a Projected Gradient Descent (PGD) attack under a white box scenario, where the attacker has complete knowledge about the model. The images demonstrate the initial input, the effect of the PGD attack, and the subsequent recovery efforts.

\begin{figure}[htbp]
  \centering
  \includegraphics[width=0.45\textwidth]{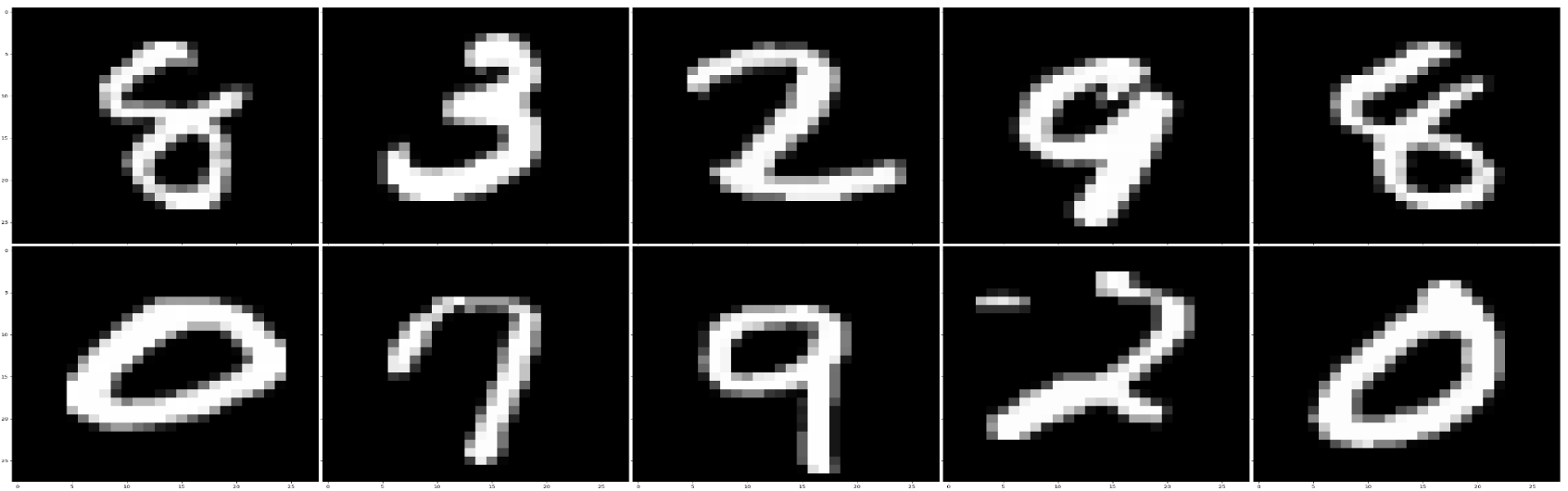}
  \caption{Input Image - White Box Attack}
  \label{fig:a}
\end{figure}

\begin{figure}[htbp]
  \centering
  \includegraphics[width=0.45\textwidth]{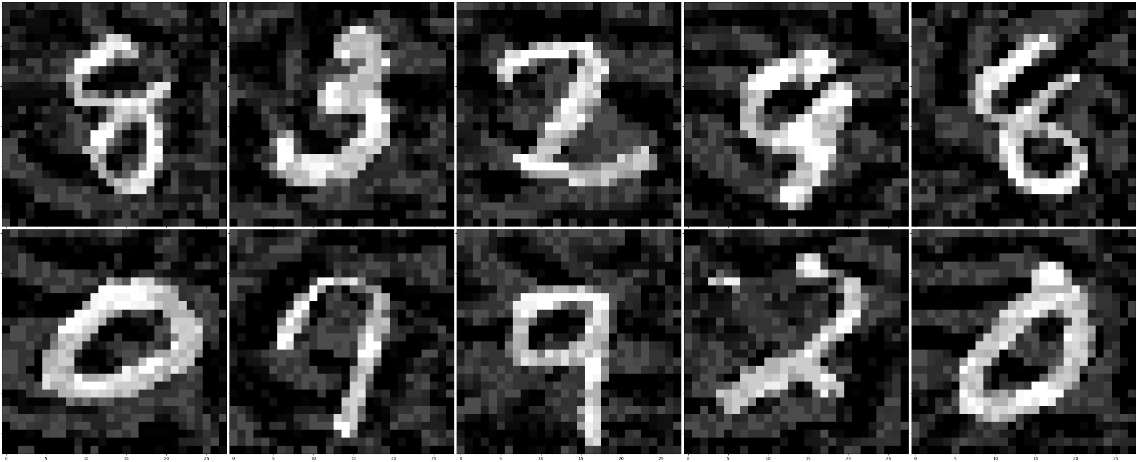}
  \caption{Attacked Image - White Box Attack}
  \label{fig:b}
\end{figure}

\begin{figure}[htbp]
  \centering
  \includegraphics[width=0.45\textwidth]{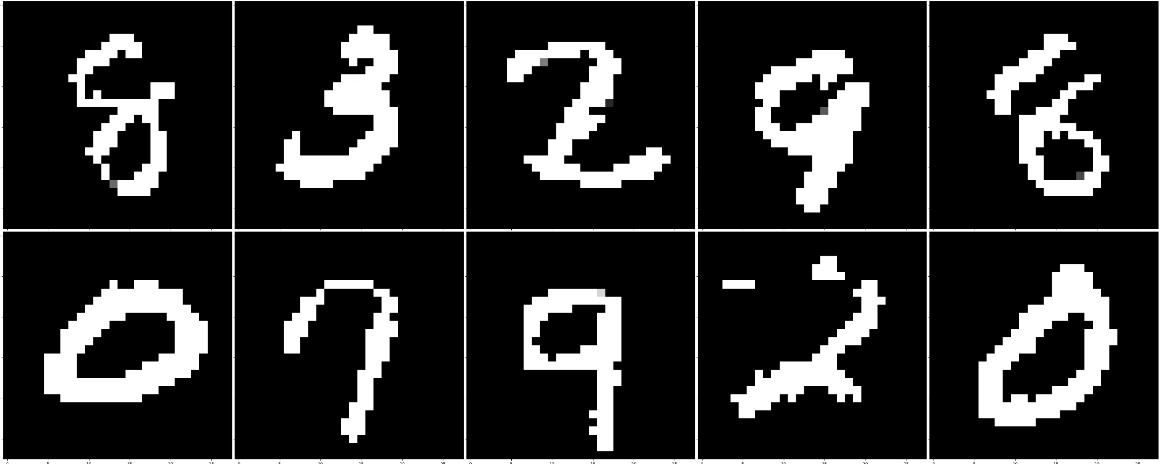}
  \caption{Recovered Image - White Box Attack}
  \label{fig:c}
\end{figure}

\begin{figure}[htbp]
  \centering
  \includegraphics[width=0.45\textwidth]{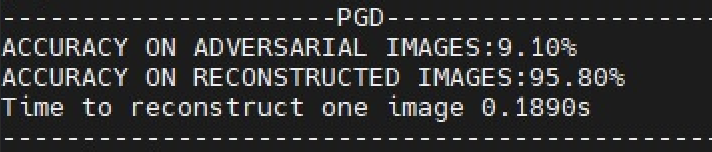}
  \caption{White Box Attack - PGD Results}
  \label{fig:wpgd}
\end{figure}

\subsection{White Box Attack - FGSM}
In this section, the Fast Gradient Sign Method (FGSM) is applied in a white box context. The series illustrates the original image, the image after a quick FGSM attack, and the effectiveness of recovery techniques in mitigating the visual impact of the attack.

\begin{figure}[htbp]
  \centering
  \includegraphics[width=0.45\textwidth]{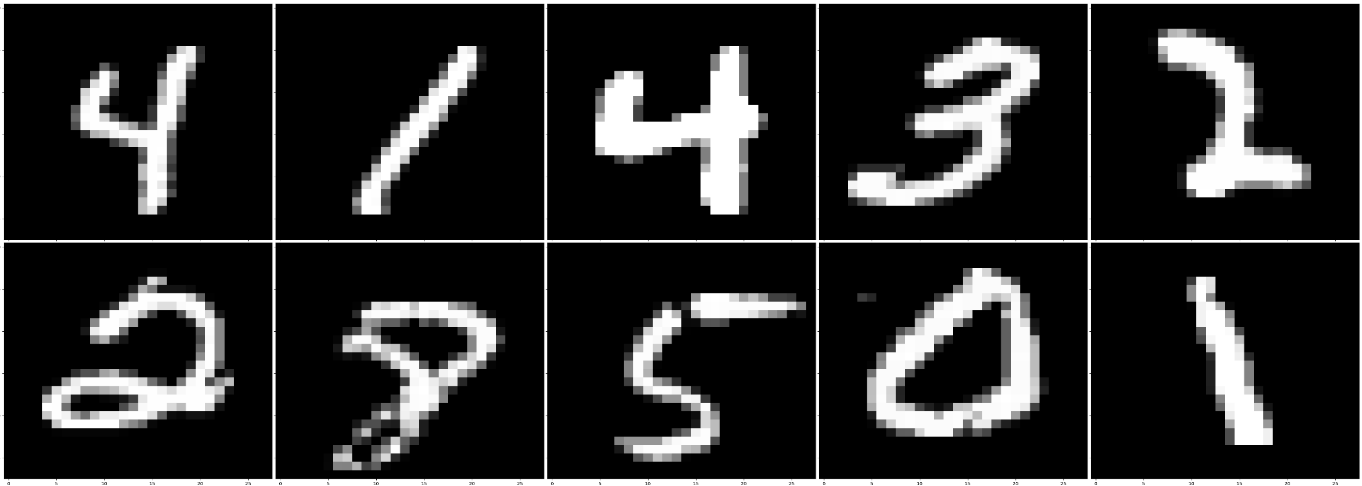}
  \caption{Input Image for White Box Attack}
  \label{fig:d}
\end{figure}

\begin{figure}[htbp]
  \centering
  \includegraphics[width=0.45\textwidth]{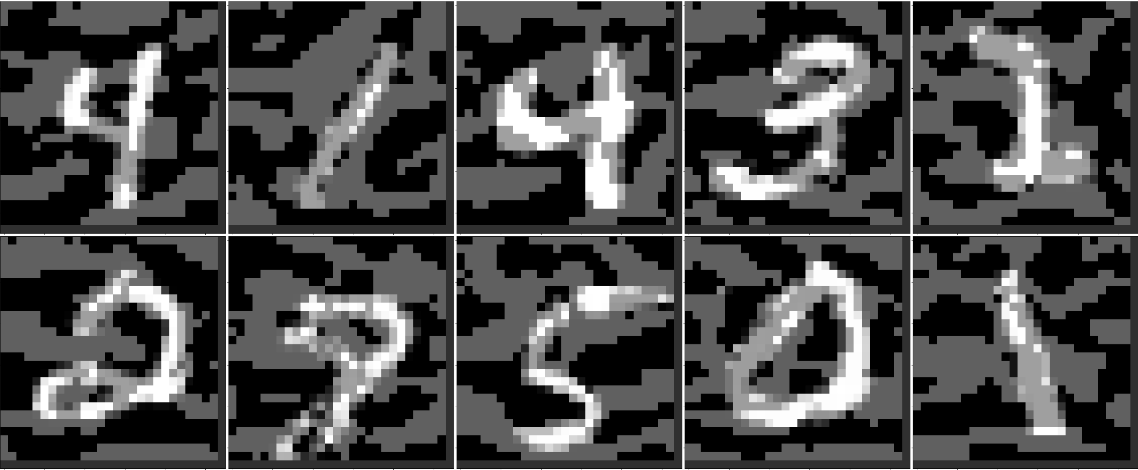}
  \caption{Attacked Image - White Box Attack}
  \label{fig:e}
\end{figure}

\begin{figure}[htbp]
  \centering
  \includegraphics[width=0.45\textwidth]{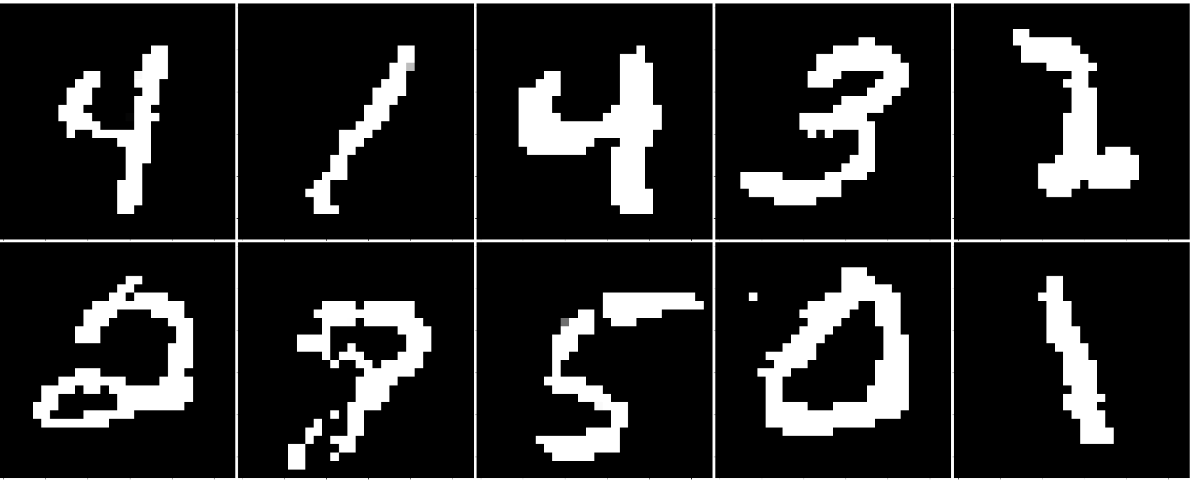}
  \caption{Recovered Image - White Box Attack}
  \label{fig:f}
\end{figure}

\begin{figure}[htbp]
  \centering
  \includegraphics[width=0.45\textwidth]{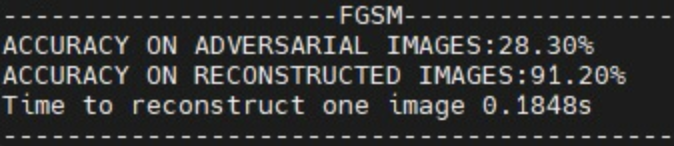}
  \caption{White Box Attack - FGSM Results}
  \label{fig:wfgsm}
\end{figure}

In the context of adversarial machine learning, black box attacks represent a critical challenge, wherein the attacker operates without direct knowledge of the model's internal structure, parameters, or training data. This form of attack is akin to probing an unknown system, relying on the observable input-output relationship to deduce potential vulnerabilities.

\subsection{Black Box Attack - PGD}

The Projected Gradient Descent (PGD) attack in a black box setting underscores the sophisticated nature of adversarial tactics. In this scenario, attackers generate perturbations based on the output of the model, iteratively refining their attack without ever seeing the underlying architecture. The images below illustrate the sequence of the PGD attack stages.

\begin{figure}[htbp]
  \centering
  \includegraphics[width=0.45\textwidth]{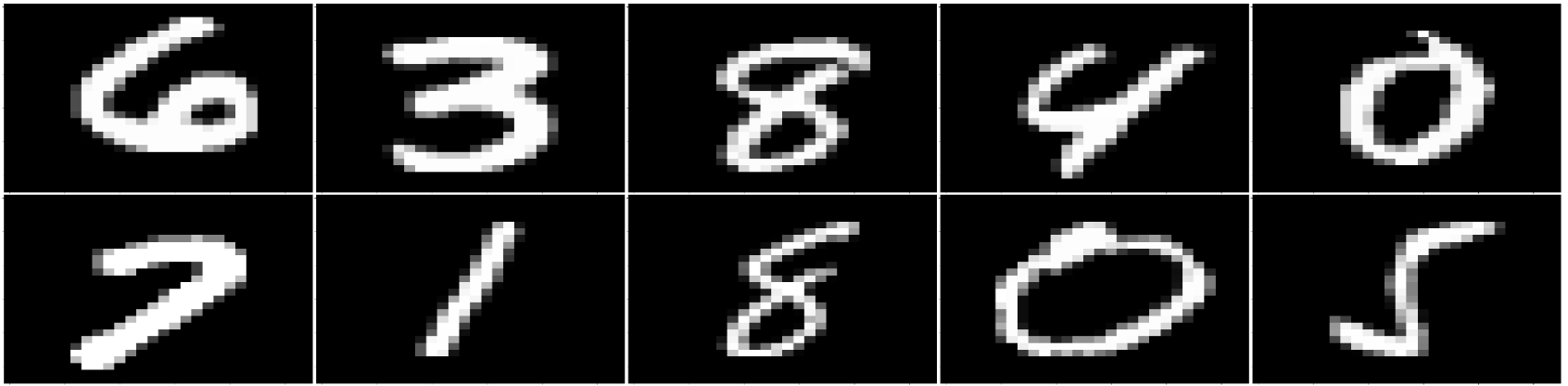}
  \caption{Input Image for Black Box Attack}
  \label{fig:g}
\end{figure}

\begin{figure}[htbp]
  \centering
  \includegraphics[width=0.45\textwidth]{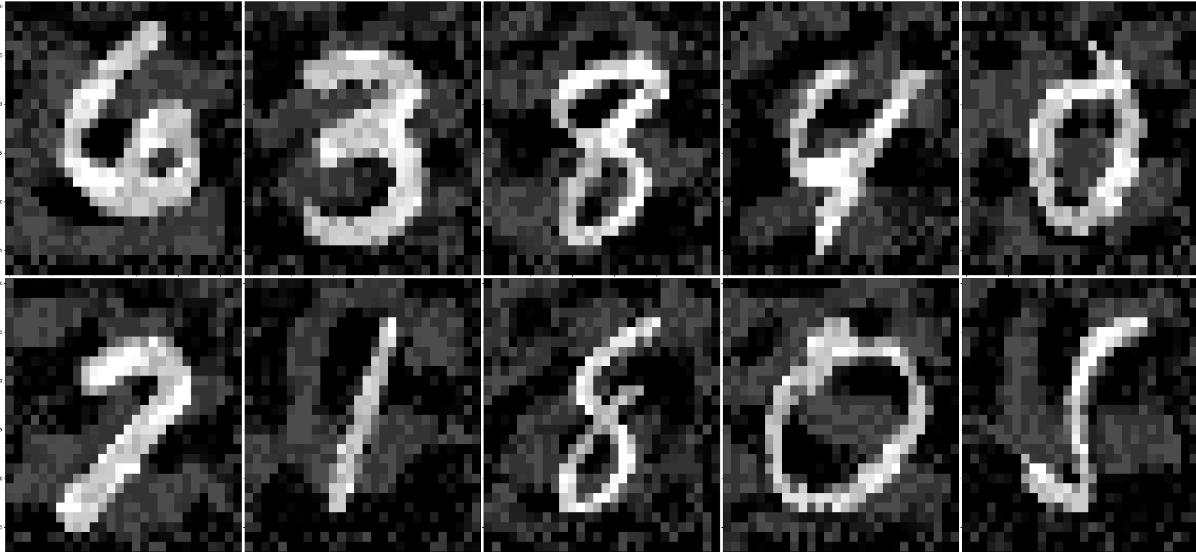}
  \caption{Attacked Image - Black Box Attack}
  \label{fig:h}
\end{figure}

\begin{figure}[htbp]
  \centering
  \includegraphics[width=0.45\textwidth]{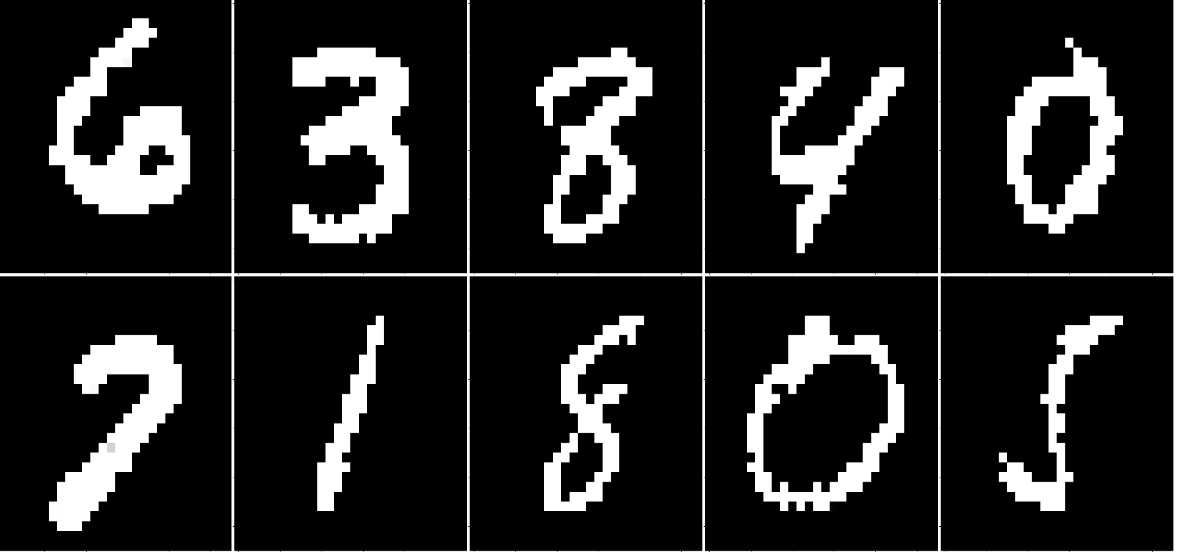}
  \caption{Recovered Image - Black Box Attack}
  \label{fig:i}
\end{figure}

\begin{figure}[htbp]
  \centering
  \includegraphics[width=0.45\textwidth]{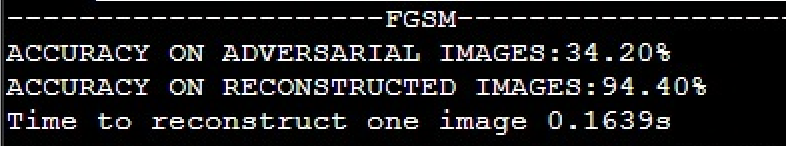}
  \caption{Black Box Attack - PGD Results}
  \label{fig:bfgsm}
\end{figure}

\subsection{Black Box Attack - FGSM}

The black box attack paradigm represents a realistic threat landscape where attackers have only external access to the machine learning model. Without the knowledge of the model's parameters or training data, attackers use the model's predictions to craft adversarial inputs. The Fast Gradient Sign Method (FGSM) is particularly noteworthy for its effectiveness in such constrained environments.

\begin{figure}[htbp]
  \centering
  \includegraphics[width=0.45\textwidth]{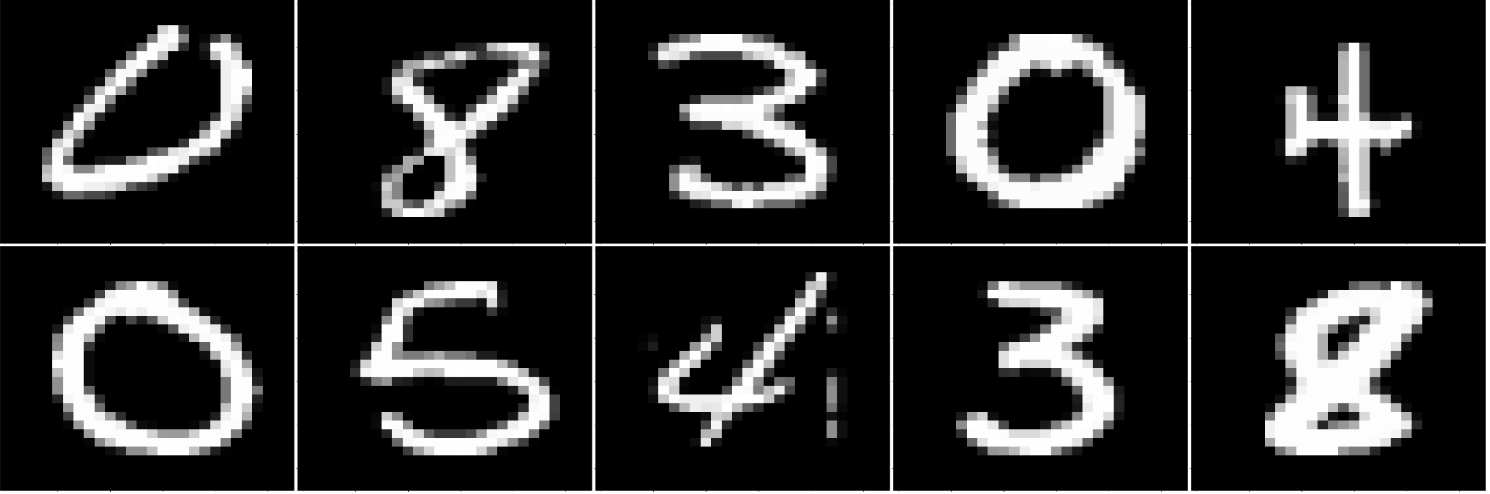}
  \caption{Input Image for Black Box Attack}
  \label{fig:j}
\end{figure}

Figure \ref{fig:j} portrays the original input used as the baseline for the attack. Despite the lack of internal model insights, FGSM leverages the gradient of the model's output with respect to the input to create a perturbed image that is likely to be misclassified.

\begin{figure}[htbp]
  \centering
  \includegraphics[width=0.45\textwidth]{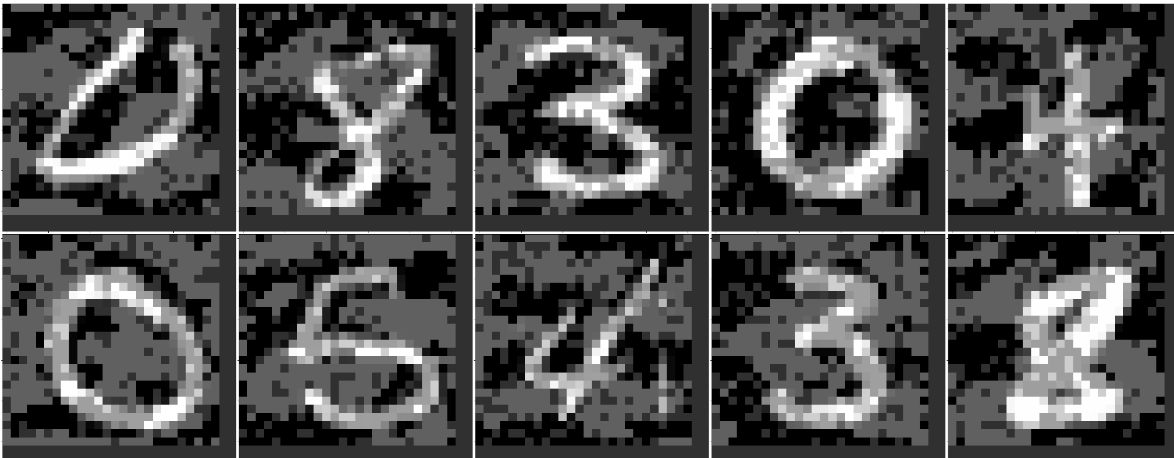}
  \caption{Attacked Image - Black Box Attack}
  \label{fig:k}
\end{figure}

The attacked image, shown in Figure \ref{fig:k}, demonstrates the subtle yet effective alterations introduced by FGSM. These alterations are generally imperceptible to the human eye but are calculated to maximise the prediction error, thereby deceiving the model into an incorrect classification.

\begin{figure}[htbp]
  \centering
  \includegraphics[width=0.45\textwidth]{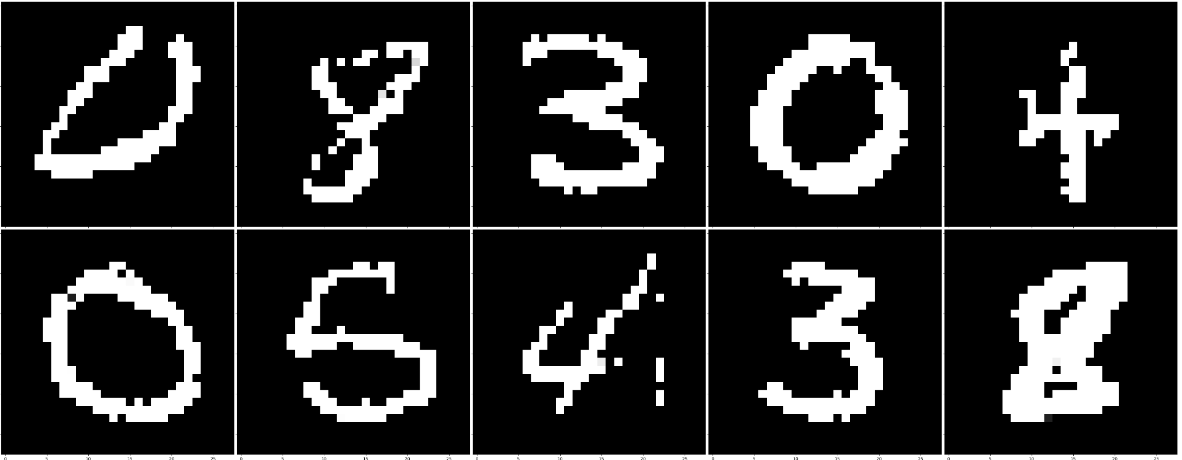}
  \caption{Recovered Image - Black Box Attack}
  \label{fig:l}
\end{figure}

\begin{figure}[htbp]
  \centering
  \includegraphics[width=0.45\textwidth]{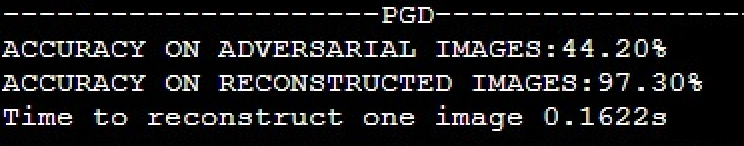}
  \caption{Black Box Attack - FGSM Results}
  \label{fig:bpgd}
\end{figure}

Figure \ref{fig:l} reveals the recovered image after employing defensive techniques against the FGSM attack. These countermeasures can include model hardening strategies such as adversarial training, input preprocessing, or applying model ensembles. The restoration of correct classification affirms the model's improved robustness and highlights the dynamic interplay between attack methodologies and defense strategies.

The FGSM attack underscores the potential vulnerabilities in machine learning models and the need for ongoing vigilance and improvement in defense mechanisms. It acts as a catalyst for the development of more secure AI systems, promoting an arms race between adversarial attacks and defensive techniques. The continued evolution of such methodologies is vital to building resilient systems capable of withstanding the ever-growing sophistication of adversarial threats.

\section{Summary}
\begin{table}[ht]
	\centering
	\label{tab:attack_performance}
	{\small 
	\begin{tabular}{|l|c|c|c|c|}
		\hline
		& \multicolumn{2}{c|}{\textbf{PGD}} & \multicolumn{2}{c|}{\textbf{FGSM}} \\ \cline{2-5} 
		\textbf{Type} & \textbf{Without} & \textbf{With} & \textbf{Without} & \textbf{With} \\ 
		& \textbf{Diffusion} & \textbf{Diffusion} & \textbf{Diffusion} & \textbf{Diffusion} \\ \hline
		White Box & 90.8 & 4.2 & 71.7 & 8.8 \\ \hline
		Black Box & 55.8 & 2.7 & 65.8 & 5.6 \\ \hline
	\end{tabular}
    \caption{Adversarial Attack Success Rates (\%)}
	}
\end{table}

The results presented in Table 1 offer a comprehensive view of the performance of two adversarial attack methods, PGD and FGSM, both with and without the implementation of diffusion techniques. Notably, the application of diffusion significantly reduces the success rates of attacks in both white box and black box scenarios, with the white box attacks showing a more pronounced decrease. This reduction in success rates underlines the effectiveness of diffusion methods in safeguarding against adversarial attacks.

In white box attacks, where attackers have full insight into the model, the PGD method without diffusion results in a high success rate of 90.8\%, indicating a substantial vulnerability. However, when diffusion is applied, the success rate plummets to 4.2\%, demonstrating the robust defense afforded by diffusion strategies. A similar pattern is observed with FGSM, where the success rate drops from 71.7\% to 8.8\% with the application of diffusion, reinforcing the technique's protective strength.

The black box attacks, characterized by the attacker's limited knowledge of the model's internals, exhibit lower success rates initially, which are further reduced with diffusion. The PGD method shows a decrease from 55.8\% to a mere 2.7\%, while FGSM shows a decrease from 65.8\% to 5.6\%. These results suggest that diffusion techniques not only enhance security against well-informed attackers but also serve as a deterrent against more general, less-informed adversarial strategies.

Overall, the data indicates that diffusion methods are not only beneficial but perhaps essential for contemporary models to withstand sophisticated adversarial attacks. Further research and development into these techniques could prove to be a vital step toward the creation of resilient AI systems in an increasingly adversarial digital landscape.



\section{Artifacts}

For practical implementation and further exploration, refer to the following example code repository, which provides a comprehensive framework for addressing the issues discussed: \href{https://github.com/Lasterminator/CPSC-8570-Network-Technologies-Security}{GitHub link}

\end{document}